\documentclass[rn,11pt]{ucl_document} 
\usepackage{graphicx}
\usepackage{natbib}

\usepackage{hyperref} 

\newlength{\halfwidth}
\newlength{\halfwidthextra}
\newlength{\temp}
\newlength{\tempa}

\begin{document}
\title{Faster Genetic Programming {GPquick} \\via multicore and Advanced Vector Extensions}
\author{
William B. Langdon (UCL)
\and
Wolfgang Banzhaf (MSU) 
}

\date{23 February 2019}
\documentnumber{19/01}

\maketitle

\pagenumbering{arabic}  
\pagestyle{plain}

\begin{abstract}
We evolve floating point Sextic polynomial populations of genetic programming
binary trees for up to a million generations.
Programs with almost
400\,000\,000 
instructions
are created by crossover.
To support unbounded Long-Term Evolution Experiment LTEE GP
we
use both SIMD parallel AVX 512 bit instructions
and 48~threads to yield 
performance of up to 139~billion GP operations per second,
139~giga GPops,
on a single 
Intel Xeon Gold 6126 2.60\,GHz
server.
\end{abstract}

{\bf keywords:}
genetic algorithms, genetic programming, 
GP,
Convergence,
Long-Term Evolution Experiment LTEE
Extended 
unbounded evolution

\section{Introduction}

A couple of years we noted~\cite{Langdon:2017:GECCO}
Rich Lenski's experiments in long term evolution
\cite{Lenski:2015:PRoySocB}
in which the BEACON team evolved bacteria for more than 70\,000
generations
and found continued beneficial adaptive mutations.
This
prompted us to ask the same question in computation based evolution.
We started to investigate
what happens if we allow artificial evolution,
specifically genetic programming~(GP) with crossover
\cite{koza:book,banzhaf:1997:book,poli08:fieldguide},
to evolve for tens of thousands, even a hundred thousand generations.
Since then new hardware has become available
and we build a new GP engine based on
Andy Singleton's GPQUICK (Section~\ref{sec:eval}).
This allows us to switch from the Boolean to the continuous domain and
run experiments of up to a million generations.
Excluding some special applications or Boolean benchmarks
based on graphics hardware (GPUs),
at up to 139 billion GP operations per second (139~giga GPops),
this appears to be the fastest single computer GP system
\cite[Tab.~3]{langdon:2013:ecgpu}.

\begin{figure} 
\centerline{\includegraphics{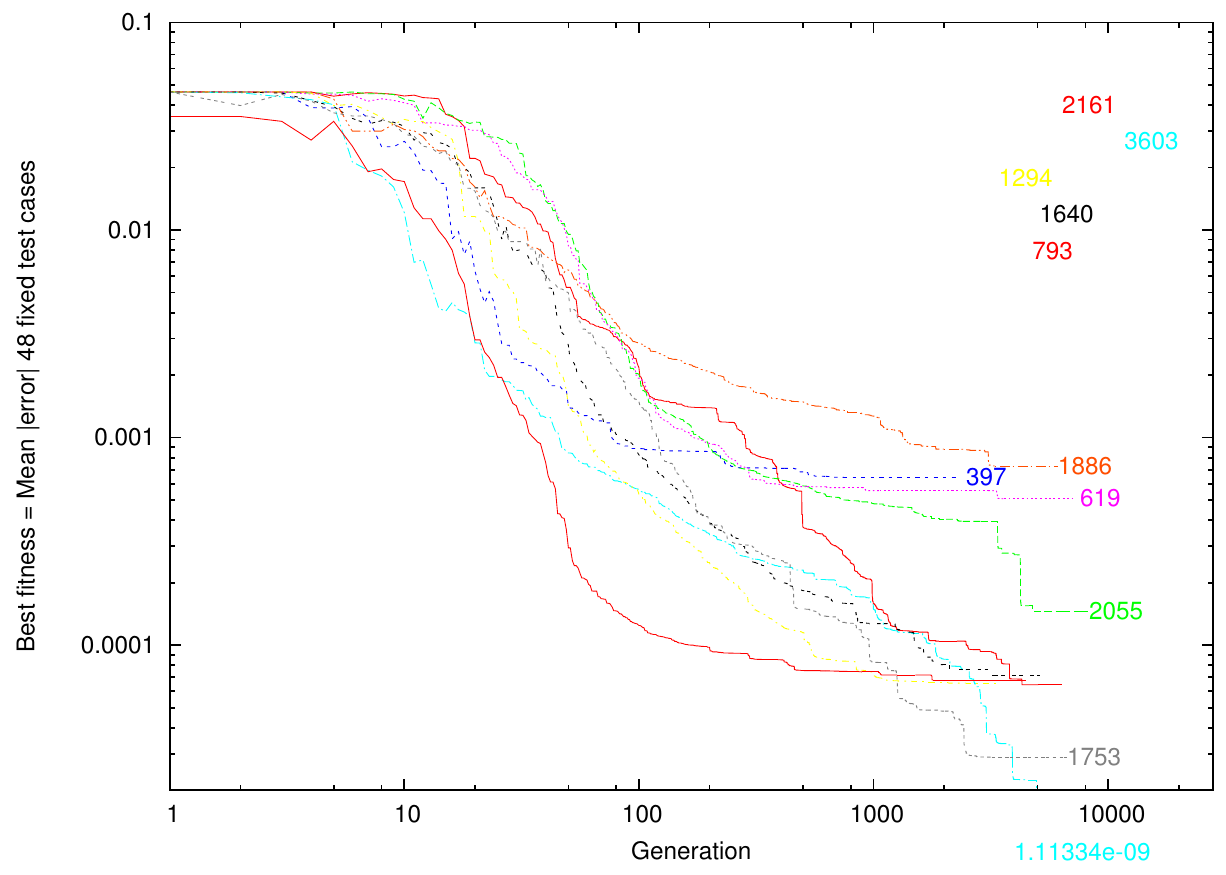}} 
\vspace*{-2ex}
\caption{\label{fig:sextic_p4000_fit}
Evolution of mean absolute error 
in ten runs of Sextic polynomial~\protect\cite{koza:book}
with population of 4000.
(Runs 
aborted after first crossover to hit
15 million node limit.)
End of run label gives
number of generations when fitness got better
(five shown at top right to avoid crowding).
}
\end{figure}

In the Boolean domain we found usually the population quickly
found a best possible answer and then retained it exactly for
thousands of generations.
Nonetheless under subtree crossover we reported interesting
population evolution with trees continuing to evolve.
Indeed we were able to report the first signs of an eventual end of
bloat due to fitness convergence of the whole population.
We can now report in the continuous domain
we do see (like the bacteria experiments)
continual innovation and improvement in fitness.
Figure~\ref{fig:sextic_p4000_fit}
show that although the rate of innovation falls
(as in Lenski's E.~Coli%
\footnote{The E.~Coli genome contains 4.6 million DNA base pairs.}
populations),
typically better solutions are found even towards the end of the runs 
and, in these runs,
there are several hundred 
or even a few thousand generations where 
sub-tree crossover between evolved parents gave a better child.

\pagebreak[4]
We are going to run GP far longer than is normally done.
Firstly in search of continual evolution but also
noting that it is sometimes not safe to extrapolate from the first
hundred or so generations.
As an example,
McPhee~\cite[sect.~1.2]{mcphee:2001:EuroGP} 
said that his earlier studies which had reported only the first 100
generations 
could not safely be extrapolated
to 3000 generations. 

\begin{figure} 
\centerline{\includegraphics[scale=0.1]{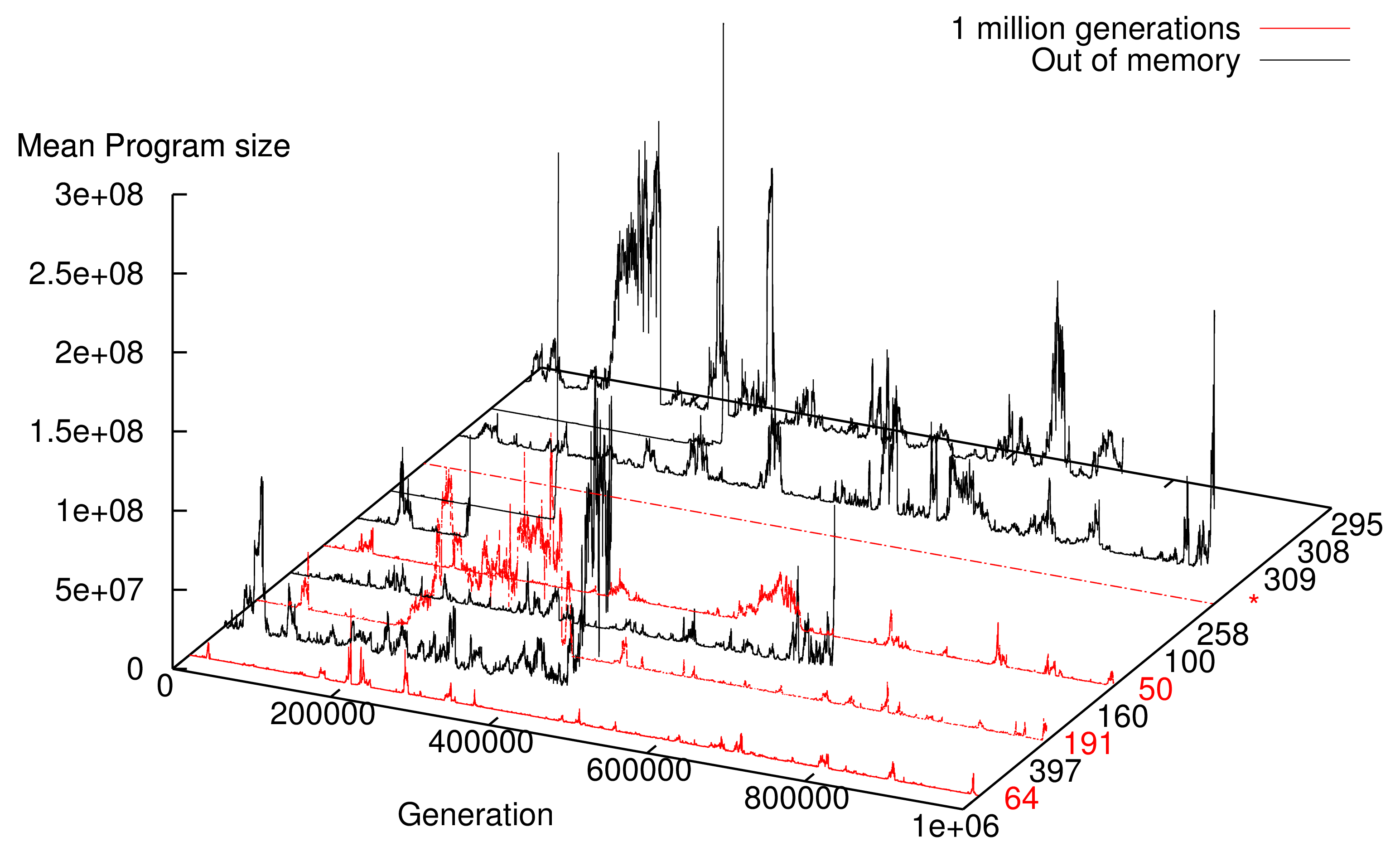}} 
\vspace*{-2ex}
\caption{
\label{fig:sextic_p48}
11 extended runs pop=48.
Numbers on right indicate size of largest tree before the run stopped
in millions of nodes.
One run~(*) converged so that more than 90\% of the trees contain
just five nodes.
Three of the other four runs that reached 1 million generations (red)
took between half a day and five days.
In all but one run~(*) we see repeated substantial bloat
($> 64$~million nodes)
and subsequent tree size collapse.
Seven runs, in black, terminated due to running out of memory
(on server with 
46GB).
}
\end{figure}

It must be admitted that without size control
we expect bloat%
\footnote{\small
\label{p.bloat}
  GP's tendency to evolve non parsimonious solutions 
  has been known since the beginning of genetic programming.
  E.g.~it is mentioned in
  Jaws \cite[page~7]{koza:book}.
  \label{p.intronsbloat}
  Walter Tackett \cite[page~45]{Tackett:1994:thesis}
  credits Andrew Singleton with
  the theory that GP bloats due to the 
  cumulative increase in 
  non-functional code, known as introns.
  The theory says these
  protect other parts of the same tree by deflecting genetic
  operations from the functional code by simply offering more
  locations for genetic operations.
  The bigger the introns, the more chance they will be hit by crossover
  and so the less chance crossover will disrupt the useful part of
  the tree.
  Hence bigger trees tend to have children with higher fitness than
  smaller trees.
  See also
  \cite{kinnear:altenberg,kinnear:angeline}.
In~\cite{Langdon:2017:GECCO} we showed prolonged evolution can produce
converged populations of 
functionally identical but genetically different trees
comprised of
the same central core of functional code next to the root node
plus a large amount of variable ineffective sacrificial code.
},
and so we need a GP system not only able to run for
$10^{6}$~generations%
\footnote{
The median run shown in 
Figure~\ref{fig:sextic_p48}
took 39 hours
(mean 62 hours).
}
but also able to process trees with well in excess of a 100~million nodes%
\footnote{
Again referring to the extended runs in Figure~\ref{fig:sextic_p48},
crossover creates highly
evolved trees  containing almost
four hundred million nodes 
These are by far the largest programs yet evolved.
}.
The new system we use is based
on Singleton's GPQuick \cite{singleton:byte,kinnear:keith,langdon:book}
but enhanced to take advantage of both
multi-core computing using pthreads
and Intel's SIMD AVX parallel floating point operations
(Section~\ref{sec:eval}).
\cite{kinnear:keith} 
say GPQuick's linearisation of the GP tree
will be hard to parallelise.
Nevertheless,
GPQUICK was rewritten to use 16~fold Intel AVX-512 instructions
to do all operations on each node in the GP tree immediately.
Leading to a single eval pass and better cache locality
but
at the expense of keeping a $T=48$ wide stack of partial results
per thread.

Although the populations never lose genetic diversity
(Koza's variety~\cite{koza:book}),
with strong tournament selection
(parent selection tournaments of size seven,
see Table~\ref{gp.details})
even the larger
populations do tend to converge to have identical fitness values.
However
100\% fitness convergence is only seen
in long runs with smaller populations (500 or 48 trees).
In contrast, in the Boolean domain~\cite{Langdon:2017:GECCO},
even in bigger populations (500)
there are many generation where 
the whole population has identical fitness
(but again variety is 100\%).

The next section describes how GPQUICK was adapted to take advantage
of Intel SIMD instructions able to process 16 floating point numbers in parallel
and Posix threads to perform crossover and fitness evaluation on
48 cores simultaneously.
Section~\ref{sex:sextic} describes the floating point benchmark
(Table~\ref{gp.details} and Figure~\ref{fig:sextic}).
Whilst Section~\ref{sec:results}
describes the evolution of fitness, size and depth
in populations of
4000, 500 and 48 trees.
It finds the earlier predictions of
sub-quadratic bloat~\cite{langdon:1999:fogp} 
and Flajolet limit
(depth $\approx\!\!\sqrt{2\pi|{\rm size}|}$
\cite{langdon:2000:fairxo})
to essentially hold.
In Section~\ref{sec:endbloat}
there is a short discussion about the continuous evolution permitted
by floating point benchmarks
before we conclude in Section~\ref{sec:conclude}.

\begin{figure}
\centerline{\includegraphics[scale=0.1]{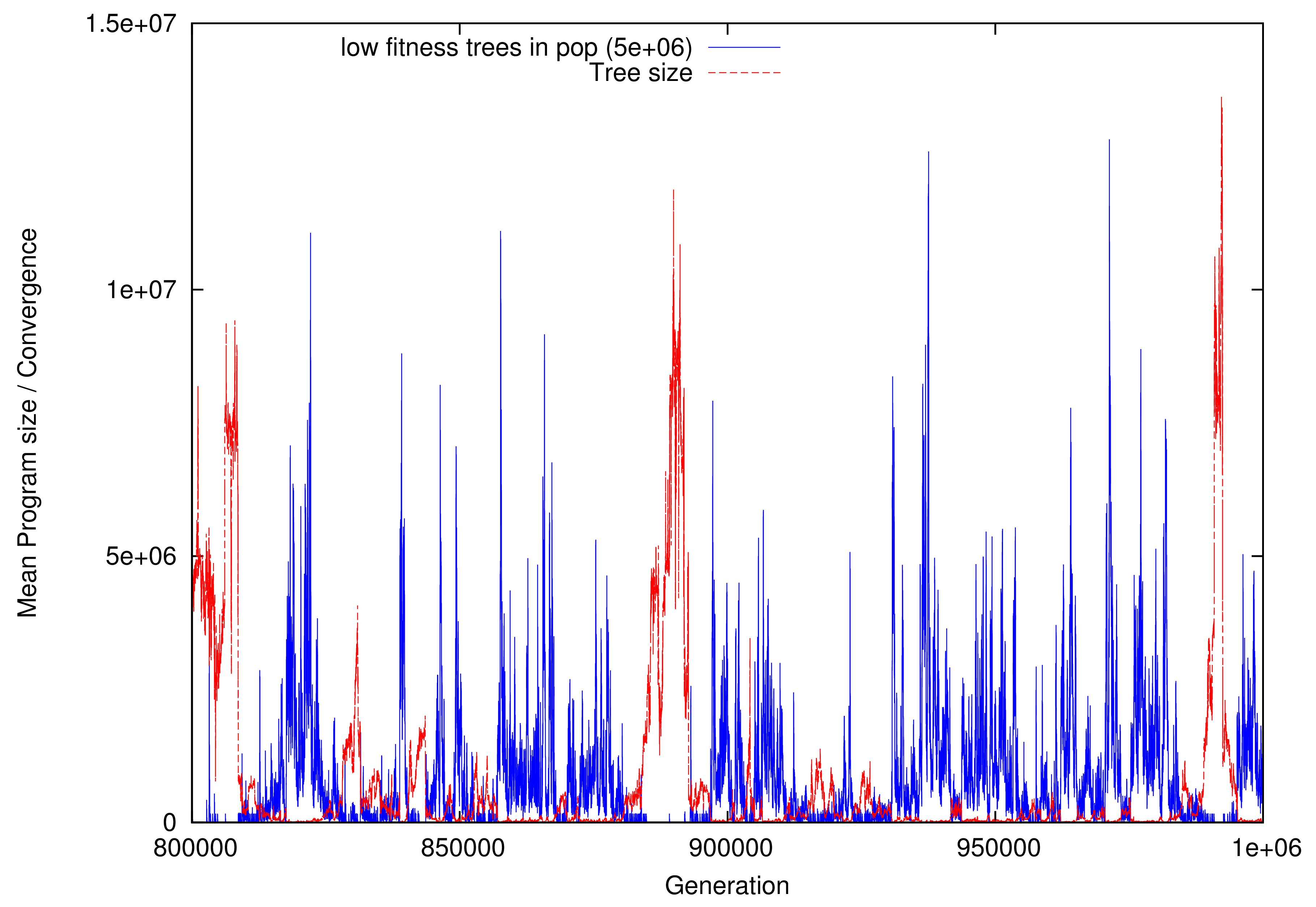}} 
\vspace*{-2ex}
\caption{
\label{fig:sextic_p48_conv}
Size and convergence in last 20\% of a typical 
extended run pop=48.
(First run ``64'' in Figure~\protect\ref{fig:sextic_p48}.)
Solid blue 
curve shows 
almost all the time there are less than 2 trees with the maximum fitness
(rescaled by $\times 5\,10^{6}$
and
smoothed over 30 generations). 
In populations with huge trees,
there are many generations where the 
whole population has identical fitness.
Without a fitness differential, tree size may rise or fall.
}
\end{figure}

\section{GPQUICK}
\label{sec:eval}
First we describe how GPQUICK is used with the Sextic polynomial
regression problem
and then how GPQUICK has been modified to run in parallel,
Sections~\ref{sec:avxeval} and~\ref{sec:pthreadseval}.

\subsection{Sextic and GPQuick}
\label{sec:qpquick}

Andy Singleton's GPQUICK~\cite{singleton:byte}
is a well established 
fast and memory efficient C++ GP framework.
In steady state mode~\cite{srgs} 
it stores GP trees in just one byte per tree node.
(Originally it supported only steady state GAs.
Ages ago a quick conversion to support generational GAs,
with separate parent and child populations
doubled this,
although 
\cite{koza:book} shows doubling is not necessary.)
The 8~bit opcode per tree node
allows GPQUICK to support a number of different functions and inputs.
Typically (as in these experiments) the remaining opcodes are used to
support about 250 fixed ephemeral random constants~\cite{poli08:fieldguide}.
In the Sextic polynomial
we have the traditional four binary floating point operations
($+$, $-$, $\times$ and protected division),
an input ($x$) and 250 constants.
The constants are chosen at random from the 2001 floating point
numbers from -1.000 to +1.000.
By chance neither end point nor 0.000 were chosen
(see Table~\ref{gp.details}).

The continuous test cases ($x$) are selected at random from the interval
-1 to +1.
At the same time the target value $y$ is calculated
(Table~\ref{gp.details}).
Since both $x$ and $y$ are stored in a text file,
\label{p.round}
there may be
slight floating point rounding errors from the
standard float$\Leftrightarrow$string conversions.

Whereas the Sextic polynomial is usually solved with 50 test cases
\cite{langdon:1999:aigp3},
since the AVX hardware naturally supports multiples of 16,
in our experiments we change this to 48 (i.e.~$3\times 16$)
(Table~\ref{gp.details}).
The multi-core servers we use each support 48 threads
and for in the longest extended runs,
we reduce the population to 48
\label{p.pop48}
(whereas in~\cite{Langdon:2017:GECCO}
the smallest population considered contained 50 trees).

\subsection{AVX GPQuick}
\label{sec:avxeval}

GPQUICK stores the GP population by flattening each
tree into a linear buffer, with the root node at the start.
To avoid heap fragmentation
the buffers are all of the same size.
Traditionally the buffer is interpreted once per test case
by multiple recursive calls to EVAL and the tree's output 
is retrieved from the return value of the outer most EVAL.
Each nested EVAL moves the instruction pointer on one position in the
tree's buffer, decodes the opcode there and calls the corresponding
function.
In the case of inputs~$x$ and constants a value
is returned via EVAL immediately,
whereas ADD, SUB, MUL and DIV will each call EVAL twice to obtain
their arguments before operating on them and returning the result.
For speed GPQUICK's FASTEVAL,
does an initial pass though the buffer and replaces all the opcodes
by the address of the corresponding function that EVAL would have called.
This expands the buffer 16 fold,
but the expanded buffer is only used during evaluation and 
can be reused by every member of the population.
Thus originally EVAL processed the tree $T+1$ times (for $T$ test cases).

The Intel AVX instructions process up to 16 floating point data
simultaneously.
The AVX version of EVAL was rewritten to take advantage of this.
Indeed as we expect trees that are far bigger than the CPU cache
($\approx\!\!16$ million bytes, depending on model),
EVAL was rewritten to process each tree's buffer only once.
This is achieved by EVAL processing all of the test cases 
for each opcode, instead of processing the whole of the tree on one
test case before moving on to the test case.
Whereas before each recursive call to EVAL returned a single floating
point value, now it has to return 48 floating point values.
This was side stepped by requiring EVAL to maintain an external stack
where each stack level contains 48 floating point values.
The AVX instructions operate directly on the top of this stack and
EVAL keeps track of which instruction is being interpreted,
where the top of the stack is,
and (with PTHREADS) which thread is running it.
Small additional arrays are used to allow fast translation from opcode
to address of eval function, and constant values.
AVX instructions are used to speed loading each constant 
into the top stack frame.
Similarly all 48 test cases~($x$) are rapidly
loaded on to the top of the stack.
However the true power comes from being able to use AVX to
process the top of the stack and the adjacent stack frame
(holding a total of 96 floats) in essentially three instructions
to give 48 floating point results.

The depth of the evaluation stack is simply the depth of the GP tree.
GPQUICK uses a fixed buffer length for everyone in the GP population.
This is fixed by the user at the start of the GP run.
Fixing the buffer size also sets  the maximum tree size.
Although in principle this only places a very weak limit on GP tree
depth,
it has been repeatedly observed 
\cite{langdon:1999:fairxo,langdon:2000:fairxo}
that evolved trees are roughly shaped
like random trees.
\label{p.Flajolet}
The mathematics of trees is well studied~\cite{segdewick:1996:aa}
in particular the depth of random binary trees
tends to a limit 
\mbox{$2\sqrt{\pi \lceil{\rm tree size}/2\rceil}$ + O(tree size$^{1/4+\epsilon}$)}
\cite[page~256]{segdewick:1996:aa}.
(See Flajolet limit in Figures~\ref{fig:p4000_french-602-3_depth},
\ref{fig:sextic_p4000_bestdepth},
\ref{fig:sextic_p500_bestdepth}
and~\ref{fig:sextic_p48_bestdepth}.)
Thus the user specified tree size limit can be readily converted into
an expected maximum depth of evolved trees.
The size of the AVX eval stack is set to this plus a suitable
allowance for random fluctuations and O(tree size$^{1/4+\epsilon}$).
Note, with very large trees,
even allowing for the number of test cases and storing floats on
the stack rather than byte sized opcodes,
the evaluation stack is considerably smaller than the genome of the
tree whose fitness it is calculating.

Although AVX allows reductions operations across a stack frame,
these are not needed until the final conversion from output to fitness
value.
However
although faster,
the reduction operations manipulate the 48 numbers in a different
order
and so may (within floating point tolerances)
produced different answers.
Since the reduction is a tiny part of the whole fitness evaluation
we decided instead 
to ensure the AVX version produces identical results
to the original system
and so the final fitness evaluation is done with a conventional for loop.

\subsection{PTHREADS GPquick}
\label{sec:pthreadseval}

The second major change to GPQUICK was
to delay fitness evaluation
so that the whole new population can have its fitness evaluated in
parallel.
(This means PTHREADS can only be applied when GPQUICK is operating in
generational mode.)
If pThreads $>0$, the population fitness evaluation is spread across
pThreads.
As trees are of different sizes,
each fitness evaluation will take different times.
Therefore which tree is evaluated by which thread is decided 
dynamically.
Due to timing variations,
in an identical run,
which tree is evaluated by which thread may be different.
However great care is taken so 
that this cannot effect the course of evolution.
(Although, for example,
where trees have identical fitness,
it can affect which is found first and 
therefore which is reported to the user).

EVAL requires a few data arrays.
These are all allocated near the start of the GP run.
Those that are read only can be shared by the threads.
Each thread requires its own instance of read-write data.
To avoid ``false sharing'',
care is taken to align read-write data on cache line boundaries
(64 bytes),
e.g.~with additional padding bytes
and {\tt ((aligned))}.
So that each thread writes to its own cache lines
and therefore these cached data are not shared with other threads.

\label{p.parallel_xo}
Surprisingly an almost doubling of speed was obtained by
also moving crossover operations to these parallel
threads.
Since crossover involves random choices of parents and subtrees
these were unchanged but instead of performing the crossover
immediately a small amount of additional information was retained and
to be read later by the threads.
This allows
the crossover to be delayed and performed in one of 48 C++ pthreads.
The results are identical but give and additional $\approx$two fold
speed up.

With intermediate sized trees,
there may be some efficiency gain by evaluating the newly created chromosome
immediately (i.e. whilst still in the hardware cache).

One thing that was less successful was to implement 
a strategy for minimising memory consumption by deleting parents
immediately all their children have been created.
(Koza's generational scheme with crossover means at most 
$M+2$ individuals need to be stored
(where $M$ is the population size).
In principle this grows to $M+2t$
(where $t$ is the number of threads).
This of course gives no saving where the population is 
the same size or smaller than
the number of threads.
Although getting the threads
to free memory immediately works,
C++ free is surprisingly slow in a multi-threaded environment
(within threads it has to wait on locks to avoid corrupting its heap).
This seriously impacted performance
(unless the time taken by EVAL dominates).
Therefore the calls to delete parents were moved back out of the threads
to the surrounding sequential code.

In principle,
even with multiple threads~$t\!\!>\!\!1$,
the efficient reduction of memory from $2M$ to $M+2t$ chromosomes
should be possible.
The number and size of all buffers is known in advance
and it would be possible to manipulate the allocation and freeing
explicitly in GPQUICK's chrome.cxx C++ source directly
(rather than using new and delete).
Thus effectively building a specialised heap 
for the population
but this has not been attempted.

\section{Experiments}
\label{sex:sextic}

\begin{figure}
\centerline{\includegraphics{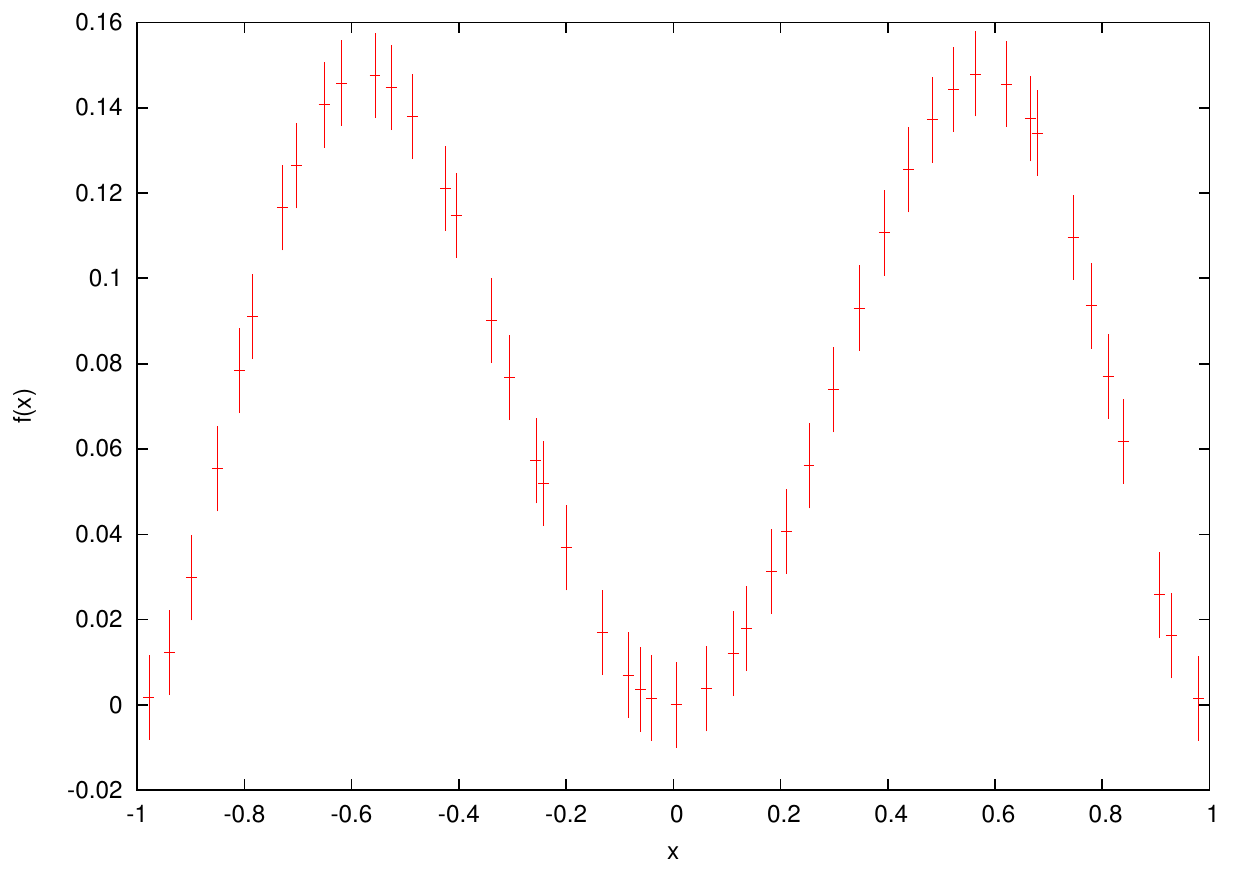}} 
\vspace*{-2ex}
\caption{\label{fig:sextic}
48 test cases for Sextic Polynomial Benchmark.
}
\end{figure}

We use the well known Sextic polynomial benchmark
\cite[Tab.~5.1]{koza:gp2}.
Briefly the task given to GP is to find an approximation to
a sixth order polynomial,
$x^{6} -2x^{4} +x^{2}$,
given only a fixed set of samples.
I.e.\
a fixed number of test cases.
For each test input $x$ we know the anticipated output $f(x)$,
see Figure~\ref{fig:sextic} and Table~\ref{gp.details}.
Of course the real point is to investigate how GP works.
How GP populations evolve over time.
In particular,
even for such a simple continuous problem,
it is possible for GP to continue to find improvements
(as Lenski's E.Coli are doing)
or, like the Boolean case~\cite{Langdon:2017:GECCO},
will the GP population get stuck early on and 
from then on never make further progress?

\begin{table}
\setlength{\temp}{\textwidth}
\settowidth{\tempa}{Fitness cases:}
\addtolength{\temp}{-\tempa}
\caption{\label{gp.details}
Long term evolution of Sextic polynomial symbolic regression binary trees
}
\begin{tabular}{@{}l@{}p{\temp}@{}}\hline
Terminal set: \rule[1ex]{0pt}{6pt} & X, 250 constants -0.995 to 0.997\\
Function set:                      & MUL ADD DIV SUB \\
Fitness cases:                     & 
48 fixed input -0.97789 to 0.979541
(randomly selected from -1.0 to +1.0 input).
\mbox{$y = xx(x\!-\!1)(x\!-\!1)(x\!+\!1)(\!x+\!1)$}
\\
Selection:  & tournament size 7 with fitness =
$\frac{1}{48}\sum_{i=1}^{48}|GP(x_i)-y_i|$
\\ 
Population: & 
Panmictic, non-elitist, generational.
\\ 
Parameters: &
Initial population (4000) ramped half and half
\cite{koza:book} depth between 2 and~6.
100\%~unbiased subtree crossover.
10\,000~generations
(stop run if any tree reaches limit $15\ 10^{6}$).
\\\hline
\end{tabular}
{\footnotesize
DIV is protected division
\tt (y!=0)? x/y : 1.0f}
\end{table}

We ran three sets of experiments.
In the first the new GP systems was set up as like the
original Sextic polynomial runs which reported phenotypic
convergence~\cite[Fig.~8.5]{langdon:1999:aigp3}.
The first group use a population of 5000,
the next 500
and the last 48.
(Section~\ref{p.pop48} above has already described a few
technical differences between these
and our earlier experiments~\cite{langdon:1999:aigp3}.)

\subsection{Crossover}

Each generation is created entirely using
Koza's two parent subtree crossover \cite{koza:book}.
(GPQuick creates one offspring per crossover.)
For simplicity and in the hope that this would make GP populations
easier to analyse,
both subtrees (i.e.~to be removed and to be inserted)
are chosen uniformly at random.
That is, we do not use Koza's bias in favour of internal nodes
(functions) at the expense of external nodes (leafs or inputs).
Instead the root node of the subtree 
(to be deleted or to be copied)
is chosen uniformly at random
from the whole of the parent tree.
This means there is more chance of subtree crossover simply moving
leaf nodes
and so many children will
differ from the root node donating parent (the mum)
by just one leaf.

As mentioned above (Section~\ref{p.parallel_xo})
once fitness evaluation has been sped up by parallel processing,
for very long trees,
producing the child is surprisingly large part of the remaining run time
and so it too can be done in parallel.
However the choice of crossover points is done in the usual way
(i.e.~in sequential code, not done in parallel)
and so remains unaffected by multithreading.
This ensures the variability introduced by multiple parallel threads
does not change the course of evolution.

\vspace*{-1ex}
\subsection{Fitness Function}
\label{sec:fitness}
\vspace*{-1ex}

The fitness of every member of every generation is calculated
using the same fitness function as 
That is, baring rounding errors (page~\pageref{p.round}),
fitness is given by the mean of the absolute difference
between the value returned by the GP tree on
each test case and the Sextic polynomial's value for the same test input
(see Table~\ref{gp.details}).
However unlike Koza,
we use tournament selection to chose both parents.

Like 
\cite[Tab.~5.1]{koza:gp2},
we also keep track of the number of 
test case where each tree is close to the target 
(i.e.~within 0.01, known as a ``hit'').
The number of hits is used for reporting the success of a GP run.
It is not used internally during a GP run.
Also our GP runs do not stop when a solution is found (48~hits)
but continue until either the 
user specified
number of generations is reached
or bloat means the GP runs out of memory.

Where needed, floating point calculations are done in a fixed order,
to avoid parallelism creating minor changes in calculated fitness,
which could quickly cause otherwise identical runs to diverge
because of implementation differences in parallel calculations.
(Also mentioned above above in Section~\ref{sec:avxeval}.)

\vspace*{-1ex}
\section{Results} 
\label{sec:results}

\vspace*{-2ex}
\subsection{Results Population 4000 trees} 
\vspace*{-1ex}

In the first set of experiments,
we use the standard population of 4000 trees.
Table~\ref{tab:p4000_french-602-3}
summarises the results of 10 runs.
In all cases GP found a reasonable approximation to the target
(the Sextic polynomial).
Indeed in all but one run (47~hits)
the best trees score 48 out of 48 possible hits.
I.e.\ they are within 0.01 on all 48 test cases.
Indeed in most cases the average error was less than
$10^{-4}$.
Figure~\ref{fig:sextic_p4000_fit}
shows that GP tends to creep up on the best match to the training data.
Typically after several thousand generations,
GP has progressively improved by more than a
thousand increasingly small steps.
(See Table~\ref{tab:p4000_french-602-3} column~3
and Figure~\ref{fig:sextic_p4000_fit}).

\begin{figure} 
\centerline{\includegraphics{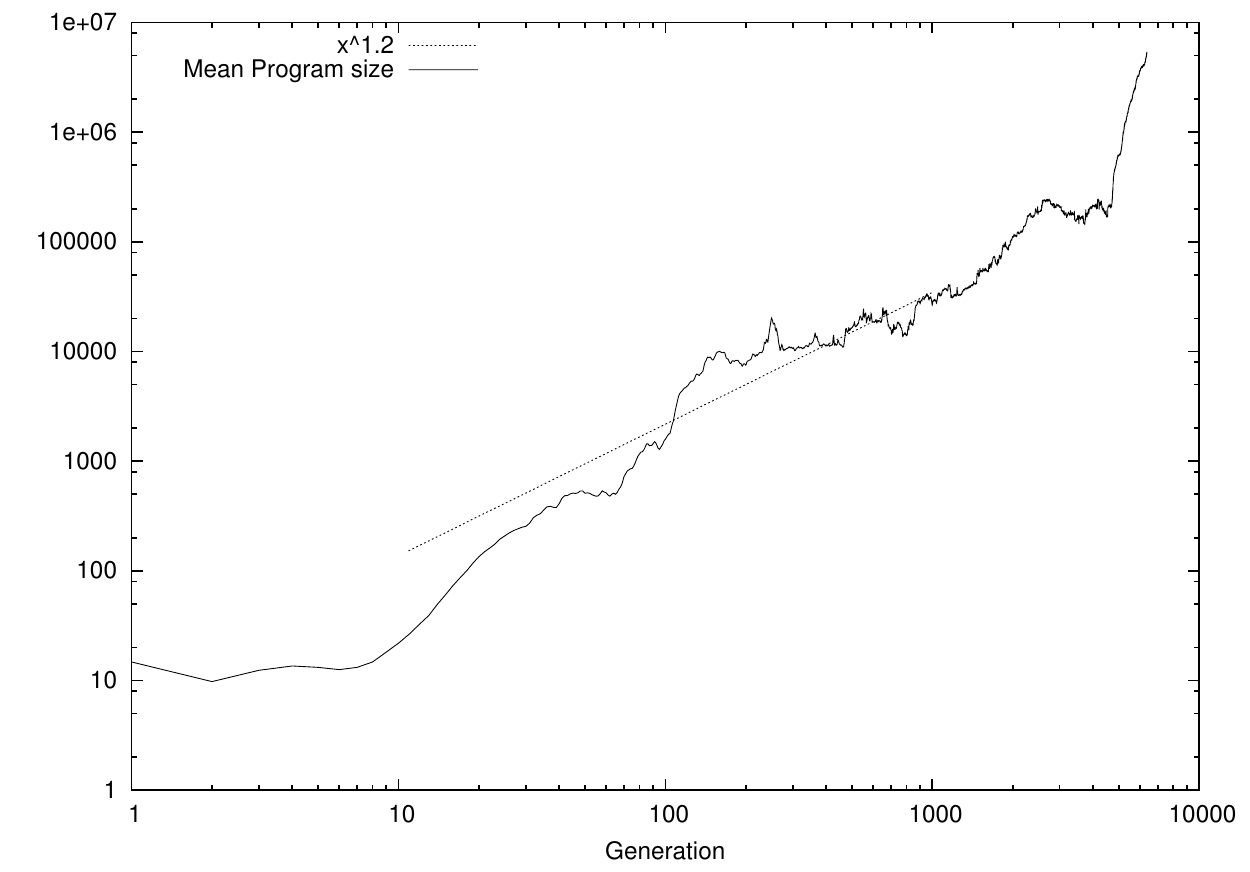}} 
\vspace*{-2ex}
\caption{\label{fig:sextic_p4000_201a_size}
Evolution of tree size in first Sextic run (population 4000).
(This run 
aborted after 6370 generations by first crossover to hit
15 million node limit.)
Straight line shows best RMS error power law fit
between generation 10 and 1000,
$y = 8.65 x^{1.2001}$
}
\end{figure}

In all cases we do see enormous increases in size.
In all ten runs with a population of 4000,
the runs are stopped as they hit the size limit
(15\,000\,000)
before reaching 10\,000 generations.
Column~5 in Table~\ref{tab:p4000_french-602-3}
gives the size (in millions) of the largest evolved tree in each run.
The log-log plot in
Figure~\ref{fig:sextic_p4000_201a_size}
shows a typical pattern of sub quadratic 
\cite{langdon:2000:quad}
increase in tree size.
The straight line, shows a power law fit.
In this run the best fit has an exponent of~1.2.
Column~6 of Table~\ref{tab:p4000_french-602-3}
shows that the best fit between generations ten and a thousand
for all 10 runs
varies between 1.1 and~1.9.

\begin{table} 
\caption{\label{tab:p4000_french-602-3}
10 Sextic polynomial runs with population 4000}
\begin{center}
\begin{tabular}{@{}rlrrrcr}
Gens &
\multicolumn{1}{@{}c@{}}{smallest error} &
\multicolumn{1}{c@{}}{impr\footnotemark} &
hits &
\multicolumn{1}{c@{}}{size$10^{6}$} &
power law &
conv\\
 6370 & 0.000064487 & 2139 & 48 & 14.329 & 1.200 & 3981 \\
 8298 & 0.000145796 & 2040 & 48 & 14.102 & 1.916 & 3982 \\
 2323 & 0.000642006 &  389 & 47 & 13.441 & 1.387 & 3995 \\
 7119 & 0.000507600 &  608 & 48 & 13.668 & 1.589 & 3997 \\
11750 & 0.000000001 & 3583 & 48 & 13.854 & 1.364 & 3989 \\
 3412 & 0.000065562 & 1277 & 48 & 14.348 & 1.625 & 3986 \\
 5106 & 0.000071289 & 1615 & 48 & 14.233 & 1.146 & 3988 \\
 6112 & 0.000728757 & 1871 & 48 & 14.500 & 1.254 & 3983 \\
 6679 & 0.000028853 & 1741 & 48 & 14.022 & 1.396 & 3998 \\
 4454 & 0.000067817 &  790 & 48 & 14.900 & 1.227 & 3997 \\

\end{tabular}
\end{center}
{
\addtocounter{footnote}{-1}
\footnotemark
Figure~\protect\ref{fig:sextic_p4000_fit}
gives number of generations which improve on their parents,
whereas here we give strictly better than anything previously
evolved.
Hence slight differences.
}
\end{table}

\begin{figure}
\centerline{\includegraphics{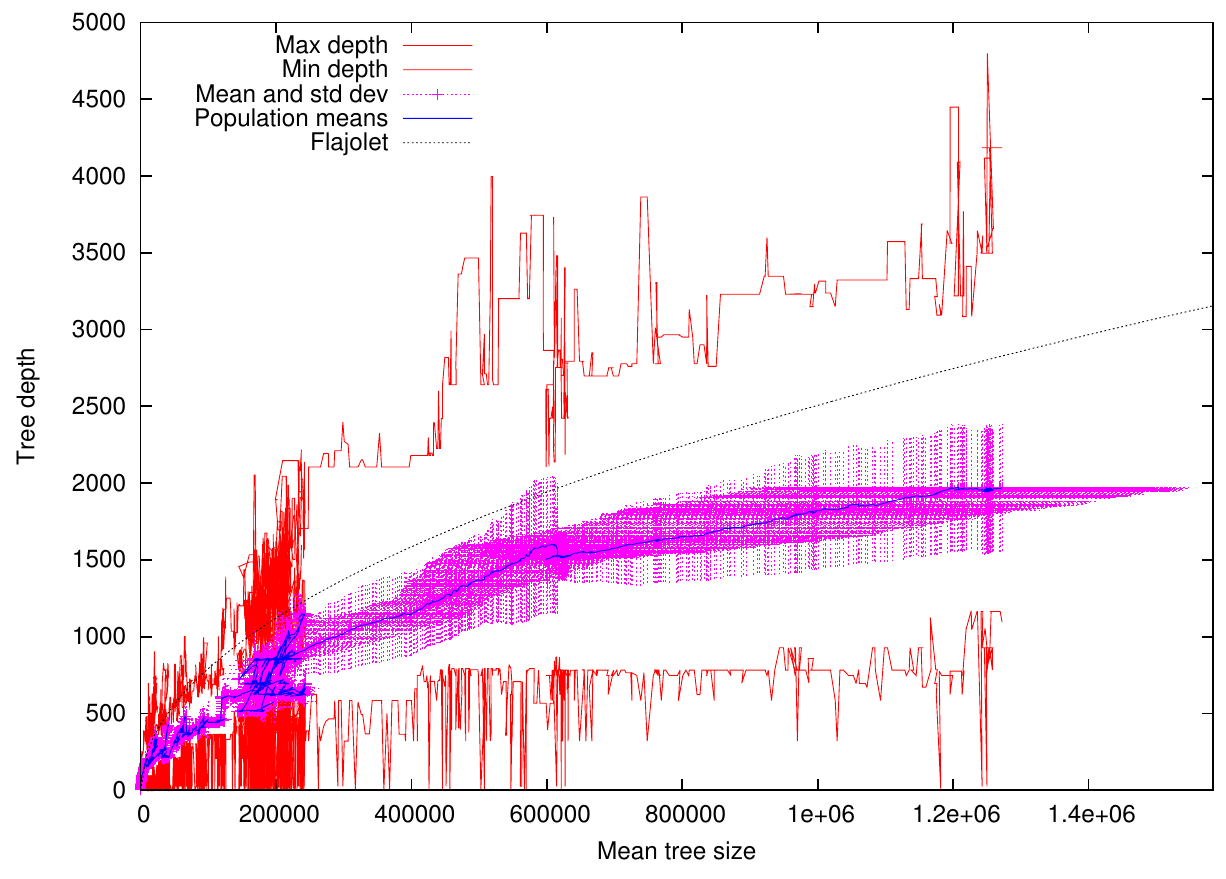}} 
\vspace*{-2ex}
\caption{\label{fig:p4000_french-602-3_depth}
Evolution of depth and size in population of 4000 trees
in typical Sextic polynomial run.
In this run,
as predicted~\protect
\cite{langdon:1999:sptfs},
average trees lie within one or two standard deviations
of random binary trees 
(Flajolet limit, depth $\approx\!\!\sqrt{2\pi|{\rm size}|}$,
\protect\cite{segdewick:1996:aa}, dotted parabola).
See also Figure~\protect\ref{fig:sextic_p4000_bestdepth}.
}
\end{figure}

\begin{figure}
\centerline{\includegraphics[scale=0.1]{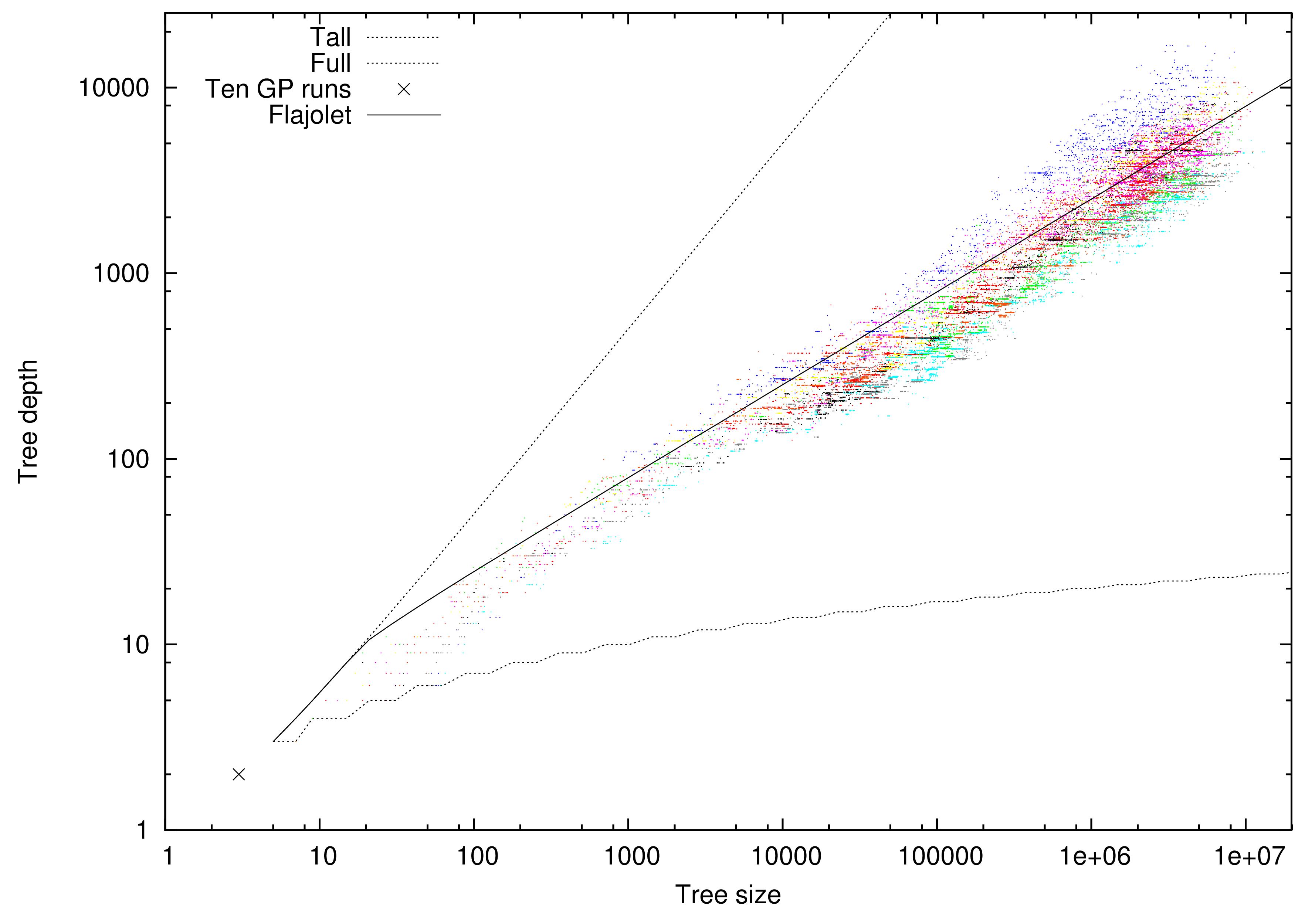}} 
\vspace*{-2ex}
\caption{\label{fig:sextic_p4000_bestdepth}
Plot of size and depth of the best individual in each generation for
10 Sextic polynomial runs with population of 4000.
Binary trees must lie between short fat trees 
(lower curve ``Full'')
and ``Tall'' stringy trees.
Most trees are randomly shaped and lie near the Flajolet limit
(depth $\approx\!\!\sqrt{2\pi|{\rm size}|}$,
solid line, note log-log scales).
Figure~\protect\ref{fig:p4000_french-602-3_depth}
shows the first run in more detail.
}
\end{figure}

As expected not only do programs evolve to be bigger but also
they increase in depth.
As described above (in Section~\ref{p.Flajolet})
highly evolved trees tend to be randomly shaped and 
so as expected tend to lie near the Flajolet limit,
depth $\approx\!\!\sqrt{2\pi|{\rm size}|}$
(see Figures~\ref{fig:p4000_french-602-3_depth}
and~\ref{fig:sextic_p4000_bestdepth}).

\begin{figure} 
\centerline{\includegraphics{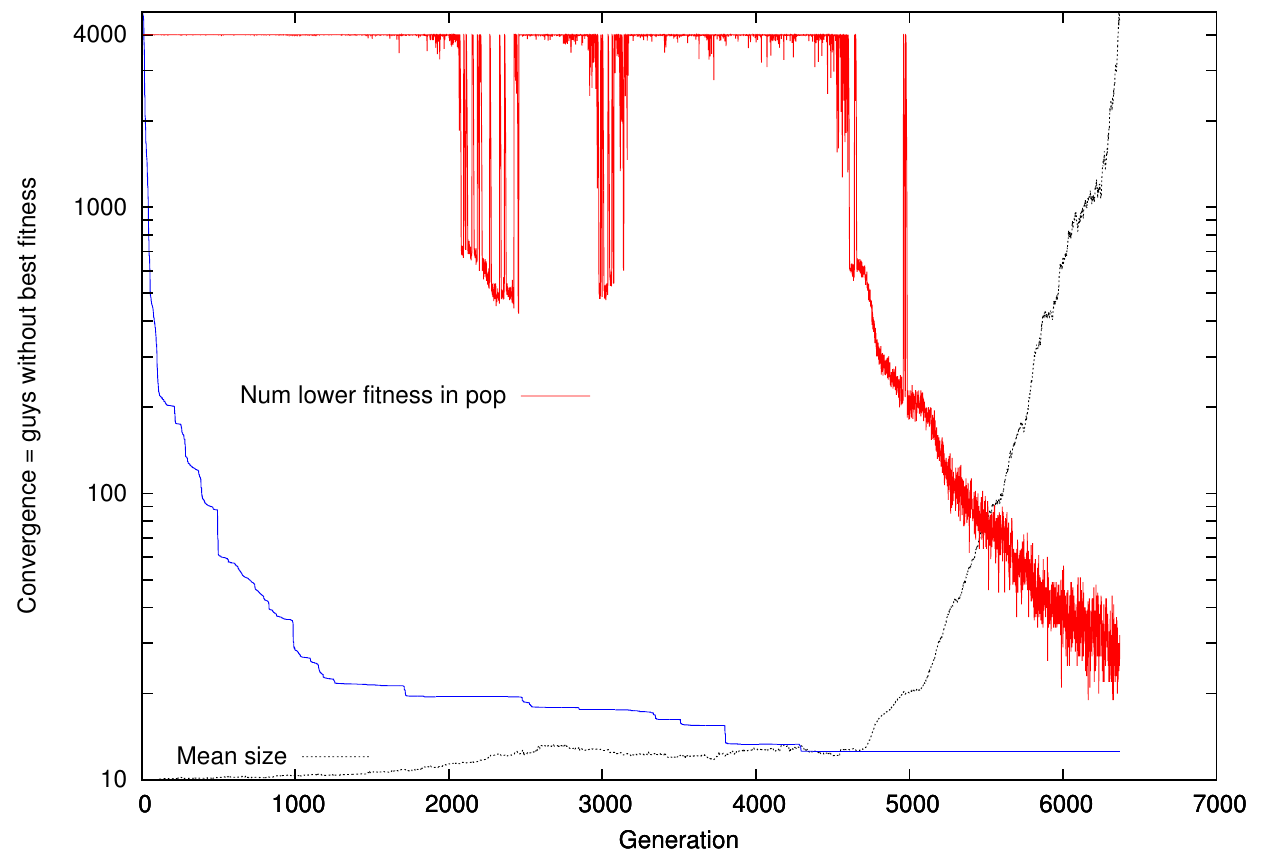}} 
\vspace*{-2ex}
\caption{\label{fig:sextic_p4000_201a_conv}
Fitness convergence in first Sextic polynomial pop=4000 run.
Perhaps because of the continual discovery of better trees
before generation 4975 and the larger population size,
although the number of tree without the best fitness falls,
unlike in the earlier Boolean problem~\protect\cite{Langdon:2017:GECCO}
it never reaches zero.
Notice tiny fitness improvement in generation 4961
resets the population for ten generations.
(Mean prog size (linear scale, dotted black)
and best fitness (log, blue)
plotted in the background.)
}
\end{figure}

In all ten runs we see some phenotypic convergence.
The
last column in Table~\ref{tab:p4000_french-602-3}
shows the peak fitness convergence.
That is, out of 4000,
the number of trees having exactly the same fitness as the best in the
population.
Typically at the start of the run
(see Figure~\ref{fig:sextic_p4000_201a_conv}),
the population contains mostly poorer trees,
but later in the run the population begins to
converge and towards the end of the run
we may see hundreds of generations where
more than 90\% of the population have identical fitness.
Under these circumstances,
even with a tournament size as high as 7,
many tournaments include potential parents with identical fitness.
These,
and hence the parents of the next generation,
are decided entirely randomly.
However,
even in the most converged population there are at least
two individuals 
with worse fitness.
(In Figure~\ref{fig:sextic_p4000_201a_conv} it is at least~19.)
As we saw with the Boolean populations~\cite{Langdon:2017:GECCO},
even this small number can be enough to drive bloat
(albeit at a lower rate).

\subsection{Results Population 500 trees}
\begin{figure}
\centerline{\includegraphics{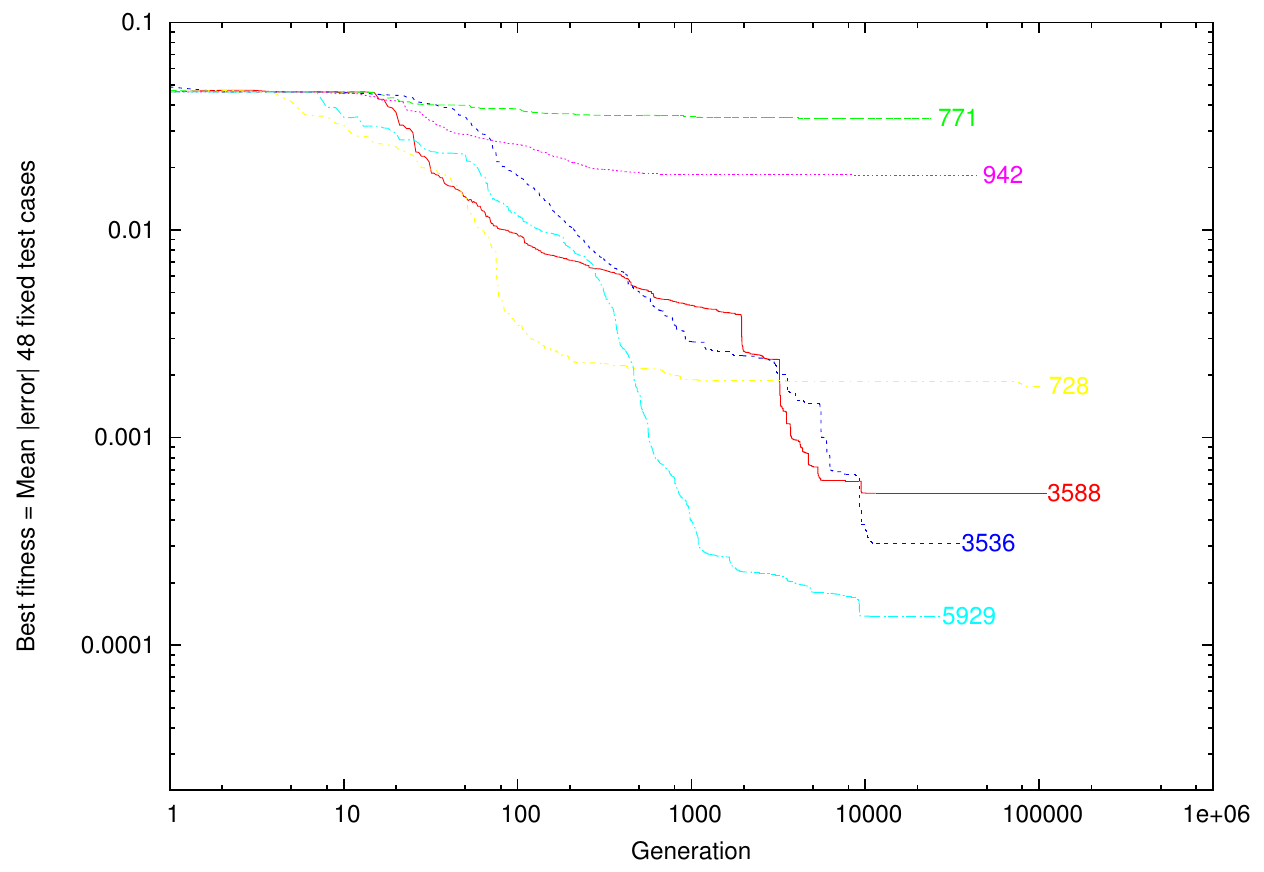}} 
\vspace*{-2ex}
\caption{\label{fig:sextic_p500_fit}
Evolution of mean absolute error 
in first six runs of Sextic polynomial~\protect\cite{koza:book}
with population of 500.
(Runs 
aborted running out of memory,
400 million node limit.)
End of run label gives
number of generations when fitness got better
}
\end{figure}

\vspace*{-1ex}
We repeated the GP runs but allowed still larger trees to evolve by
splitting the available memory between fewer trees
by reducing the population from 4000 to 500.
Figure~\ref{fig:sextic_p500_fit}
shows the evolution of the best fitness with the reduced population
size.
Notice two runs do not really solve the problem
getting less than half the test cases
(see ``hits'' column in Table~\ref{tab:p500_french-602-3}).
Nonetheless in all cases evolution continues to make
progress and each GP runs finds several hundred or more small improvements
(third column in Table~\ref{tab:p500_french-602-3}).

\vspace{-1ex}

\begin{figure}
\centerline{\includegraphics{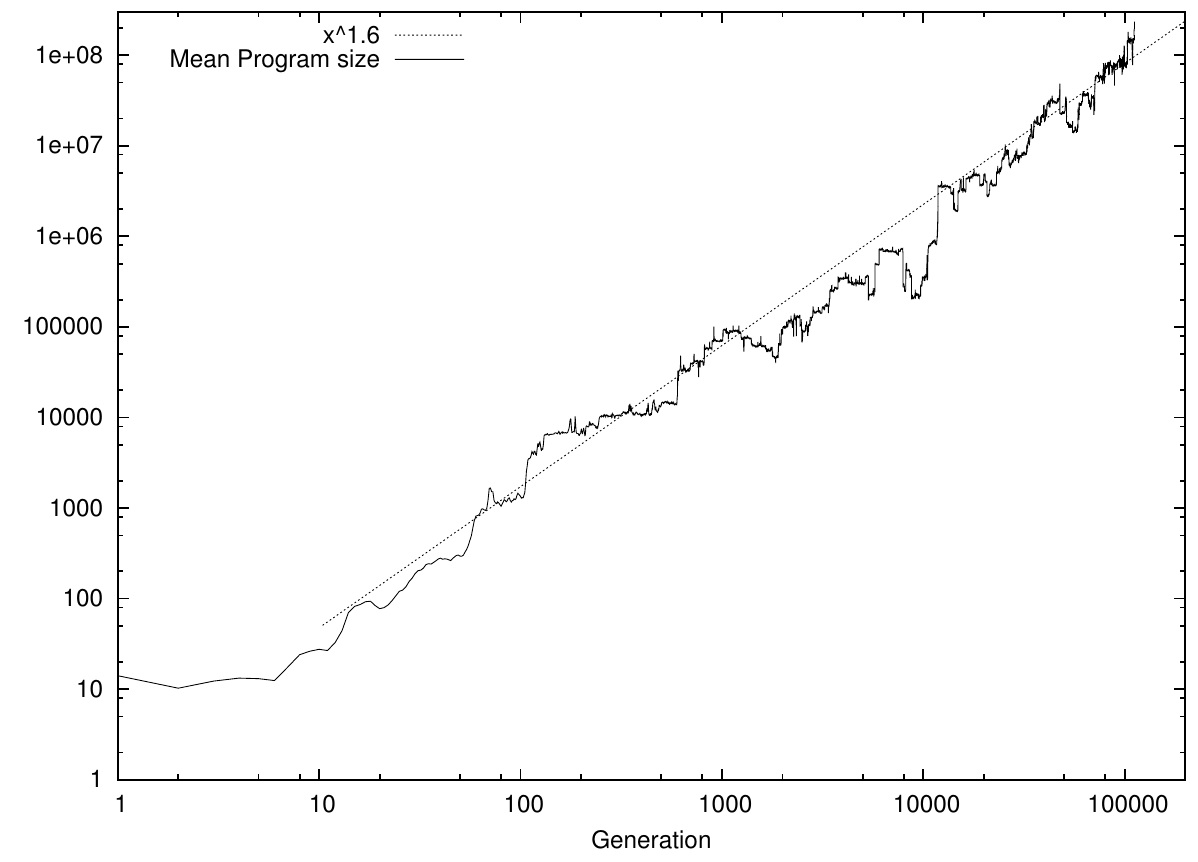}} 
\vspace*{-2ex}
\caption{\label{fig:sextic_p500_201a_size}
Evolution of tree size in first Sextic run (population 500).
(This run 
aborted after 111\,582 generations by first crossover to hit
400 million node limit.)
Straight line shows best RMS error power law fit
between generation 10 and 1000,
$y = 1.3 x^{1.56}$
}
\end{figure}

Since we have deliberately extended the space available to the GP trees,
it is no surprise that the trees grow even bigger than before
(column~5 in Table~\ref{tab:p500_french-602-3}).
Again the bloat is approximately a power law.
Although in one unsuccessful run we do see a power law exponent greater
than 2,
mostly growth is at a similar (sub-quadratic rate) as with the bigger
population runs
(1.4--2.2 v 1.1--1.9, column~6 in
Table~\ref{tab:p4000_french-602-3} (pop~4000)).
Figure~\ref{fig:sextic_p500_201a_size}
shows as expected
we again see sub-quadratic growth in
tree size
between generations 10 and 1000.
In fact the power law fit ($<$2.0),
in the  first GP run with a population of 500,
seems to extrapolate well,
even though the population starts to converge
in later generations
(see Figure~\ref{fig:sextic_p500_201a_conv}).

\begin{table} 
\caption{\label{tab:p500_french-602-3}
6 Sextic polynomial runs with population 500}
\begin{center}
\begin{tabular}{@{}rlrrrcr}
Gens &
\multicolumn{1}{@{}c@{}}{smallest error} &
\multicolumn{1}{c@{}}{impr\footnotemark} &
hits &
\multicolumn{1}{c@{}}{size$10^{6}$} &
\multicolumn{1}{c@{}}{power law} &
conv\\
111582 & 0.000538248 & 3545 & 47 & 399.594 & 1.558 &  500 \\
23937 & 0.034313100 &  757 & 18 & 202.439 & 1.736 &  500 \\
35783 & 0.000307189 & 3484 & 48 & 227.488 & 1.436 &  500 \\
43356 & 0.018373600 &  929 & 22 & 267.416 & 2.181 &  500 \\
27713 & 0.000137976 & 5852 & 48 & 327.253 & 1.928 &  500 \\
103953 & 0.001765590 &  664 & 48 & 230.106 & 1.408 &  500 \\

\end{tabular}
\end{center}
{
\addtocounter{footnote}{-1}
\footnotemark
Figure~\protect\ref{fig:sextic_p500_fit}
gives number of generations which improve on their parents,
whereas here we give strictly better than anything previously
evolved.
Hence slight differences.
}
\end{table}

\vspace{-1ex}

\begin{figure}
\centerline{\includegraphics[scale=0.1]{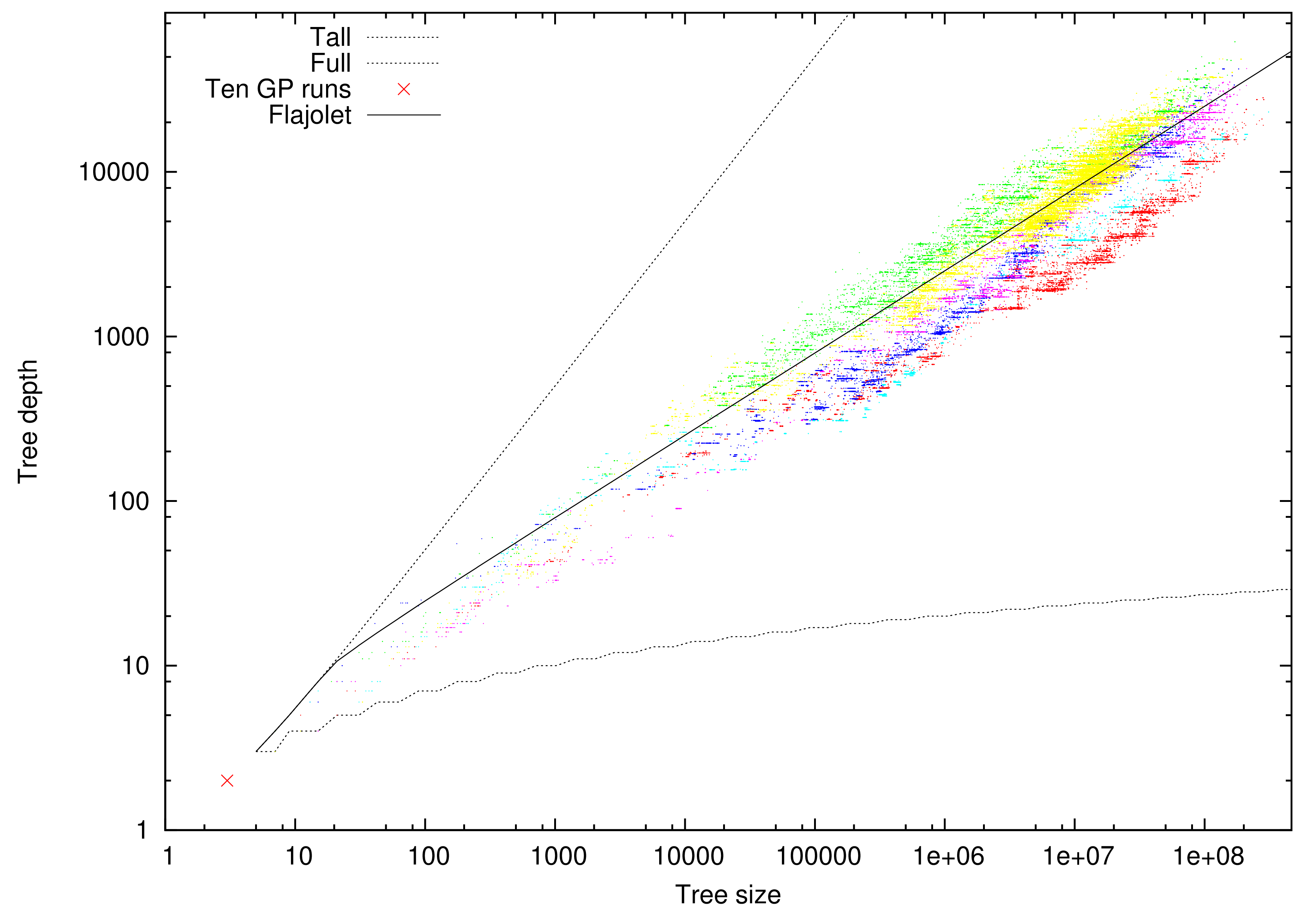}} 
\vspace*{-2ex}
\caption{\label{fig:sextic_p500_bestdepth}
Plot of size and depth of the best individual in each generation for
six Sextic polynomial runs with population of 500.
Binary trees must lie between short fat trees 
(lower curve ``Full'')
and ``Tall'' stringy trees.
Most trees are randomly shaped and lie near the Flajolet limit
(depth $\approx\!\!\sqrt{2\pi|{\rm size}|}$,
solid line, note log-log scales).
}
\end{figure}

\begin{figure}
\centerline{\includegraphics[scale=0.1]{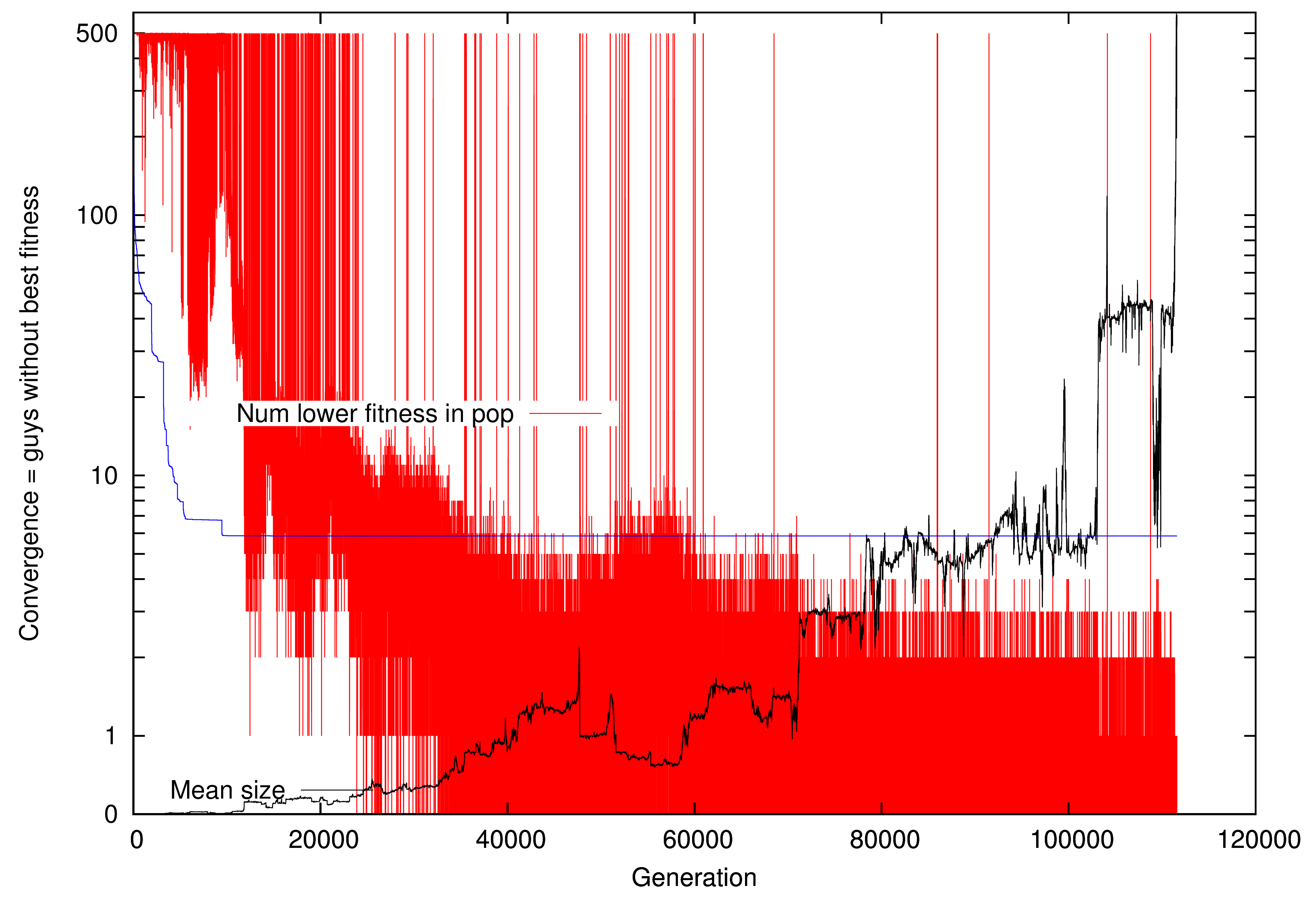}} 
\vspace*{-2ex}
\caption{\label{fig:sextic_p500_201a_conv}
Fitness convergence in first Sextic polynomial pop=500 run.
In 
30\% of this run,
the whole population has identical fitness
(y=0).
Mean prog size (linear scale, black)
and best fitness (log, blue)
plotted in the background.
}
\end{figure}

Again randomly shaped trees evolve.
Figure~\ref{fig:sextic_p500_bestdepth}
shows the relations ship between depth and size of
the best trees in the population.
As expected in all six runs it
lies near the Flajolet limit,
depth $\approx\!\!\sqrt{2\pi|{\rm size}|}$,
for large binary trees.

\vspace{-1ex}

Unlike with the large populations,
all the runs with populations of 500 trees showed
some cases of complete fitness convergence
(last column in Table~\ref{tab:p500_french-602-3} is 500).
Figure~\ref{fig:sextic_p500_201a_conv}
shows
fitness convergence in first Sextic polynomial pop=500 run.
Unlike with larger population (cf.\ Figure~\ref{fig:sextic_p4000_201a_conv}),
in this run,
the whole population has identical fitness
33\,143 times
(30\% of the run).
If we concentrate upon 
the last fitness improvement
in generation 108\,763
(2819 before the end of the run).
This new improved Sextic polynomial performance
takes over the whole population in half a dozen generations.
(Shown in the plot (Figure~\ref{fig:sextic_p500_201a_conv})
as the rightmost thin vertical red line.)
However it fails to totally dominate the population in 861 
(31\%)
of the remaining
generations.
Even though the mean number of lower fitness children
is less than one
(0.38)
it is not zero,
and this (given nearly three thousand generations)
is still enough to double the average size of the trees.

\subsection{Results Population 48 trees}
In the final experiments
the population was reduced still further to allow still larger trees
to be evolved.
Whereas in the Boolean experiments~\cite{Langdon:2017:GECCO}
we reduced the population to 50,
this was before we had access to servers with 48 cores.
Therefore these smallest population runs
were run with a population of 48,
since this should readily map well to the available Intel mult-core servers.

\begin{figure}
\centerline{\includegraphics{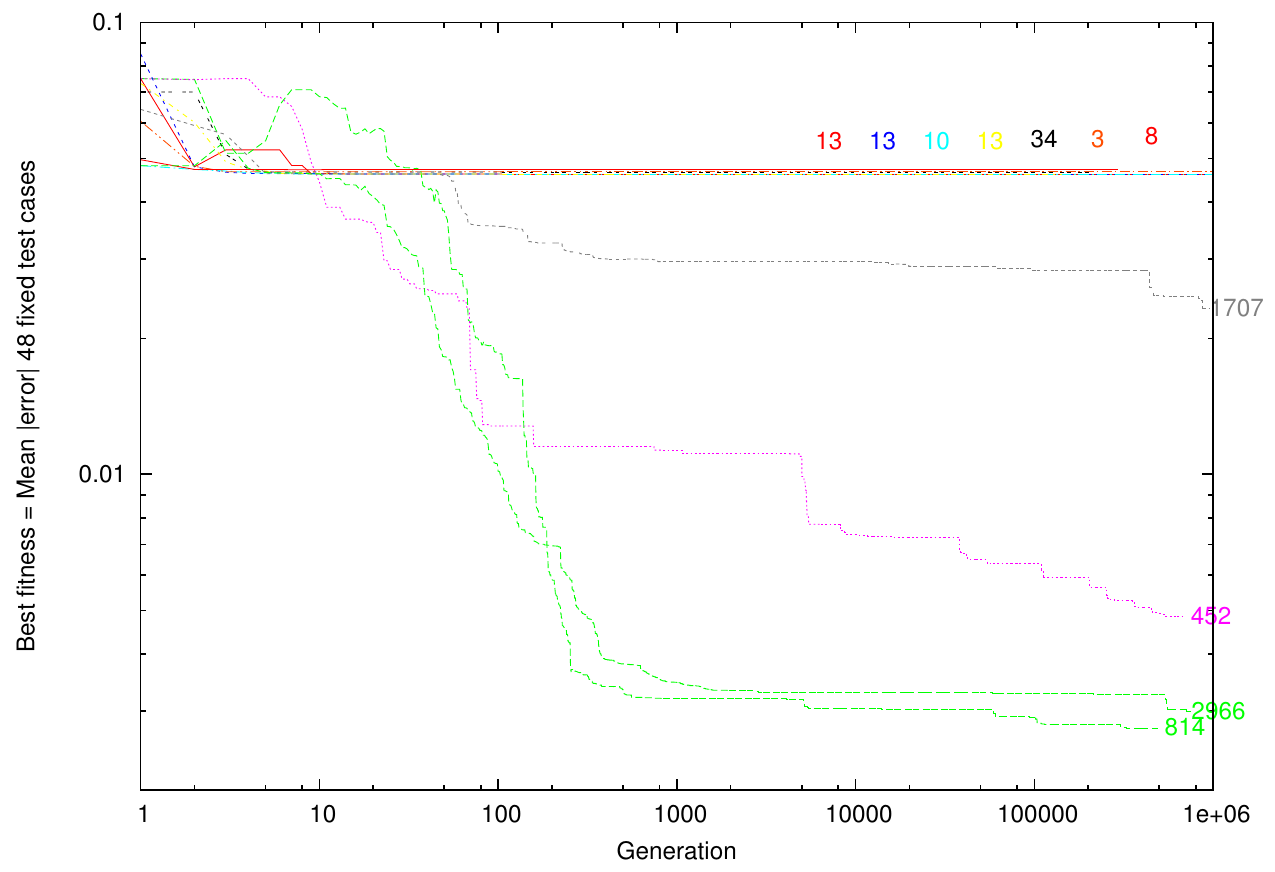}} 
\vspace*{-2ex}
\caption{\label{fig:sextic_p48_fit}
Evolution of mean absolute error 
in 11 runs of Sextic polynomial~\protect\cite{koza:book}
with population of 48.
(Up to a million generation or 
aborted on running out of memory,
500 million node limit.)
End of run label gives
number of generations when fitness got better.
(Seven shown at top right to avoid crowding.)
}
\end{figure}

With the small population,
none of the runs solve the problems.
Indeed only three runs got close on 40 or more test cases
(see Figure~\ref{fig:sextic_p48_fit}
and Table~\ref{tab:p48_french-602-3}).
Of the remaining eight,
only one finds a large number of fitness improvements.
Seven runs contain only between 3 and 30 generations 
containing fitness improvement,
column~3 in Table~\ref{tab:p48_french-602-3}.
In three of these,
the population gets trapped at trees with just three nodes
which evaluate to constants 0.0626506, 0.069169 and 0.0830508.
Although eventually the population does eventually escape
and large trees evolve by the end of the run.
Except for these three runs, all the other runs contain
populations where every member of the population has identical
fitness.
Therefore their maximum convergence is 48
(see last column in Table~\ref{tab:p48_french-602-3}).

\begin{table} 
\caption{\label{tab:p48_french-602-3}
11 Sextic polynomial runs with population of 48}
\begin{center}
\begin{tabular}{@{}rlrrrcr}
Gens &
\multicolumn{1}{@{}c@{}}{smallest error} &
\multicolumn{1}{c@{}}{impr\footnotemark} &
hits &
\multicolumn{1}{c@{}}{size$10^{6}$} &
power law &
conv\\
1000000 & 0.046215700 &   11 & 16 & 63.920 & 1.633 &   48 \\
491618 & 0.002748230 &  745 & 46 & 396.576 & 2.060 &   48 \\
1000000 & 0.046215700 &    7 & 13 & 190.654 & 1.448 &   48 \\
689414 & 0.004857990 &  448 & 40 & 159.949 & 1.260 &   48 \\
1000000 & 0.046215700 &    8 & 14 & 50.365 & 1.701 &   48 \\
143251 & 0.046215700 &   11 & 14 & 99.541 & 1.672 &   48 \\
212528 & 0.046650600 &   30 & 14 & 257.766 &    na &   42 \\
1000000 & 0.046730800 &    3 & 14 & 0.000 &    na &   42 \\
958147 & 0.023259300 & 1683 & 18 & 308.958 & 1.791 &   48 \\
294098 & 0.047174400 &    3 & 12 & 308.121 &    na &   43 \\
757830 & 0.002985980 & 2921 & 44 & 294.821 & 1.320 &   48 \\

\end{tabular}
\end{center}
{
\addtocounter{footnote}{-1}
\footnotemark
Figure~\protect\ref{fig:sextic_p48_fit}
gives number of generations which improve on their parents,
whereas here we give strictly better than anything previously
evolved.
Hence slight differences.
}
\end{table}

\begin{figure}
\centerline{\includegraphics[scale=0.1]{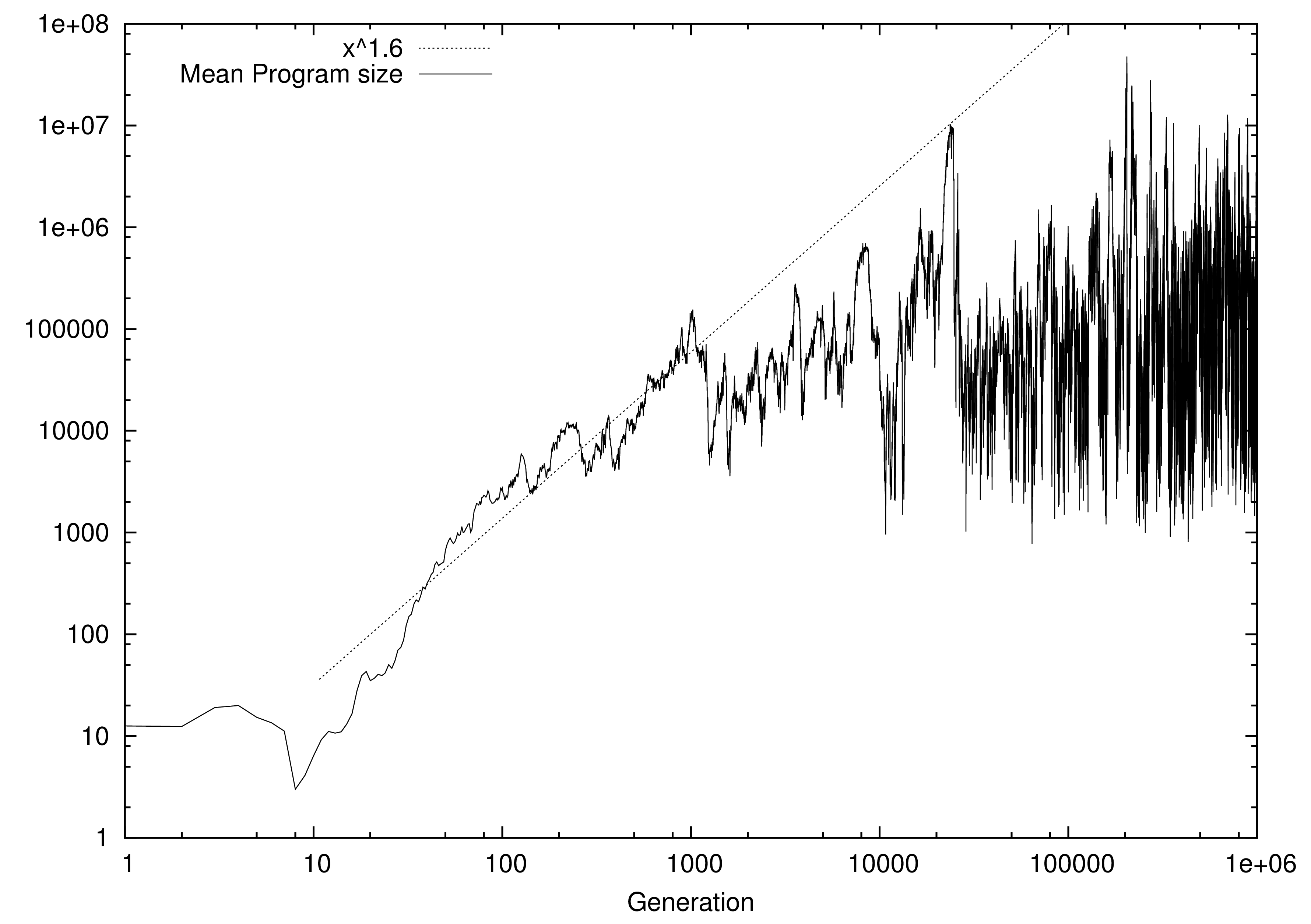}} 
\vspace*{-2ex}
\caption{\label{fig:sextic_p48_201a_size}
Evolution of tree size in first Sextic run with a population of~48.
This run 
ran for a million generations.
Straight line shows best RMS error power law fit
between generation 10 and 1000,
\mbox{$y = 0.75 x^{1.63}$}
}
\end{figure}

\begin{figure}
\centerline{\includegraphics[scale=0.1]{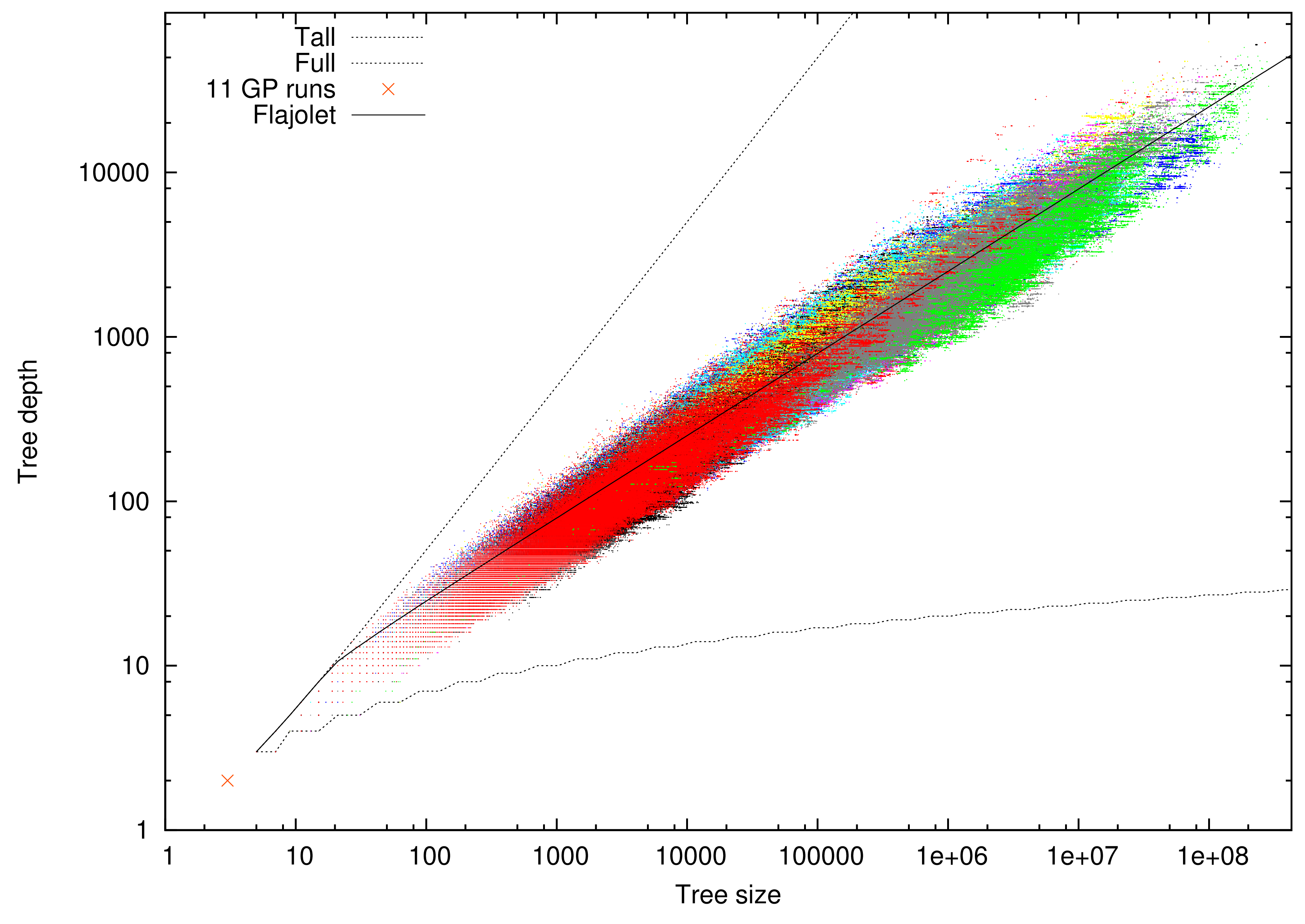}} 
\vspace*{-2ex}
\caption{\label{fig:sextic_p48_bestdepth}
Plot of size and depth of the best individual in each generation for
eleven Sextic polynomial runs with population of 48.
}
\end{figure}

As expected,
as with larger populations,
the evolved highly binary trees
are again approximately the same shape as random trees.
See Figure~\ref{fig:sextic_p48_bestdepth}.

\begin{figure}
\centerline{\includegraphics[scale=0.1]{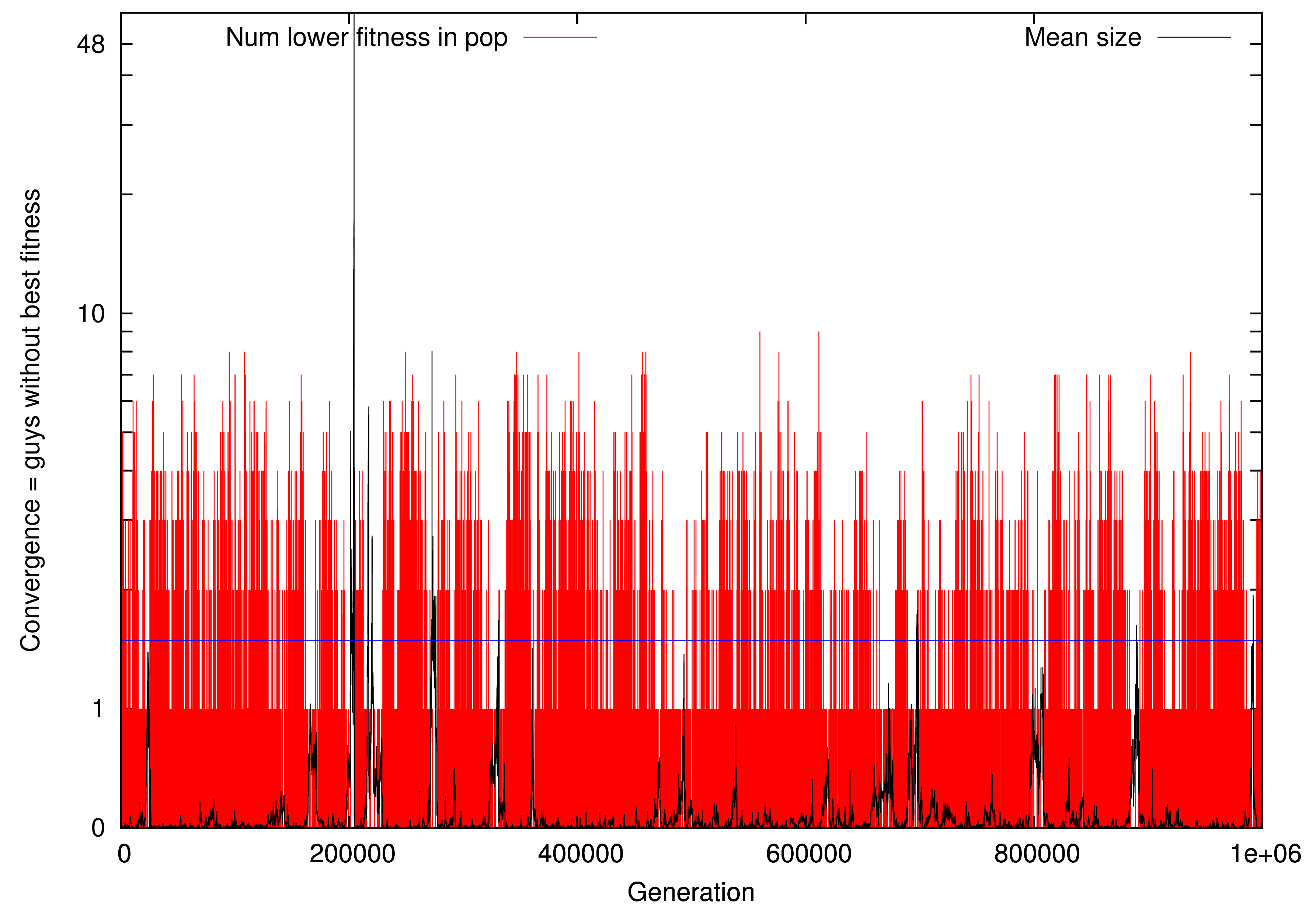}} 
\vspace*{-2ex}
\caption{\label{fig:sextic_p48_201a_conv}
Fitness convergence in first Sextic polynomial run with population of 48 trees
After generation 19 
the best tree in every generation has a fitness of 
-0.0462157 (4 hits) 
(tree returns 0.0769947 
regardless of test case,
cf.\ Figure~\protect\ref{fig:sextic}).
In 
90\% of this run,
the whole population has identical fitness
(y=0).
(Mean number of poor fitness children is 0.1315.)
Mean prog size (linear scale, black)
and best fitness (log, blue).
}
\end{figure}

Figure~\ref{fig:sextic_p48_201a_conv}
shows for almost the whole run the best fitness in the population 
is fixed but once trees get big enough
further size changes are essentially random.
Notice fitness depends only on the sum of the absolute difference
between the value returned by the GP tree and the target value.
In particular the ``hits'' is only used for reporting.
The best fitness found in this run is given by
robust trees which always return a midpoint value
(cf.\ Figure~\protect\ref{fig:sextic})
which only passes close to four test points.
Trees which closely matched more test points
were discovered in the first nineteen generation of this run.
However,
in terms of fitness, they scored worse than a constant and so went extinct.

\begin{figure}
\centerline{\includegraphics[scale=0.1]{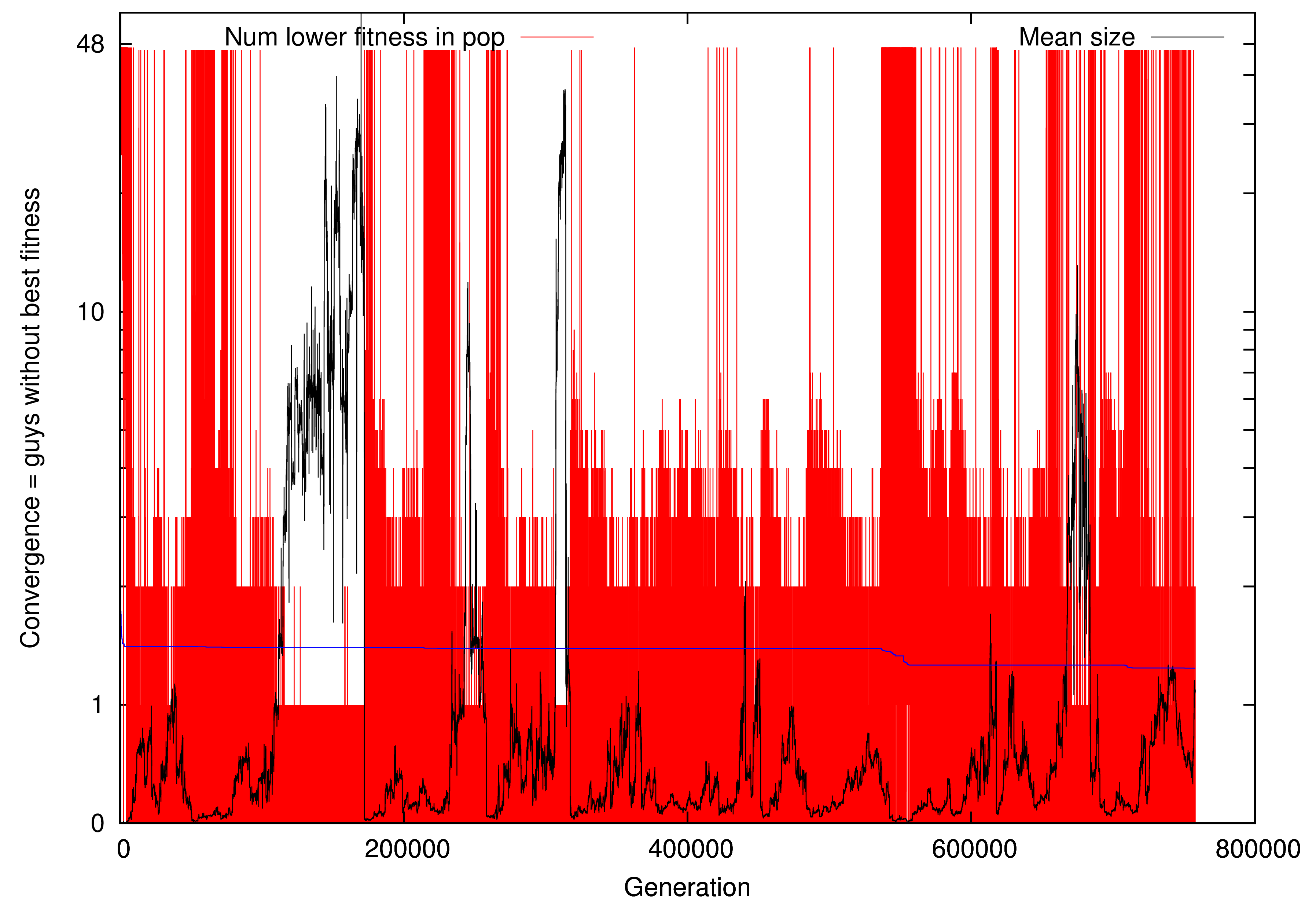}} 
\vspace*{-2ex}
\caption{\label{fig:sextic_p48_232a_conv}
Fitness convergence in last Sextic polynomial run with population of 48 trees.
(Run 295 in Figure~\protect\ref{fig:sextic_p48}.)
In this run
67\% of generations all 48 trees have identical fitness.
The last improvement is found in generation
752\,071
(5759 generation before the end of the run)
in a tree of 13\,196\,331 nodes.
It takes over the whole population in 4~generations.
However in 
21\% of the remaining generations it does not totally dominate
(mean number trees with lower fitness 0.394 per generation).
Mean prog size (linear scale, black)
and best fitness (log, blue).
}
\end{figure}

\pagebreak[4]
\section{Is there a Limit to Evolution?}
\label{sec:endbloat}

In the Sextic Polynomial experiments with larger populations
there is no hint of either evolution of fitness or bloat 
totally stopping.
In the smaller populations,
it is both possible to run evolution for longer
and to allow trees to bloat to be even larger.
Four of the eleven extended runs reached a million generations
but in the remaining seven, bloat ran into memory limits and
halted the run.
Only in one run did we see anti-bloat,
in which the population converged in a few generation on a 
small high fitness tree which crossover was able to replicate 
across a million generations.
Interestingly two other runs found similar solution but
after thousands of generations crossover found bloated version of
them.

In the binary 6-Mux Boolean problem~\cite{Langdon:2017:GECCO}
there are only 65 different fitness values.
Therefore the number of fitness improvements is very limited.
An end to bloat was found.
By which we mean
it was possible for trees to grow so large that
crossover was unable to disrupt the important part of their
calculation 
next to the root node
and many generations were evolved
where everyone had identical fitness.
This lead to random selection and random fluctuations
in tree size.
(I.e.~enormous trees but without 
a tendency for progressive end less growth.)

This did not happen here.
Even in some of the smallest Sextic polynomials runs,
we are still seeing innovation in the second half of the run,
with tiny fitness improvements being created by crossover between
enormous parents.
Also we are still slightly short of total fitness convergence.
Even with populations containing Sextic polynomial
trees of hundreds of millions of nodes,
crossover can still be disruptive and
frequently even tiny populations can contain a tree of lower fitness.
This is sufficient to provide some pressure
(over thousands of generations)
for tree size to increase on average.

{\em Can bloat continue forever?}
It is still difficult to be definitive in our answer.
We have seen cases where it does not
and of course there are plenty of engineering techniques to prevent bloat.
But we see other cases where crossover over thousands of generations
can create an innovative child which allows bloat into a converged 
population of small tress.
Perhaps more interestingly,
we see crossover finding fitness improvement 
in bloated trees after many thousand of generations.

It is still an open question in continuous domains
if,
given sufficient memory and computational resources,
bloat will always stifle innovation
so completely that crossover will always only reproduce children of
exactly the same fitness for long enough that the lack of selection
pressure 
\cite{Langdon:1997:bloatWSC2}
will in turn stifle bloat.

\section{Conclusions}
\label{sec:conclude}

The availability of mult-core SIMD capable hardware
has allowed us to push GP performance on single computers with
floating point problems to that
previously only approached with sub-machine code GP
operating in discreet domains
\cite{poli:1999:aigp3,poli:2000:22par}.
This in turn has allowed GP runs far longer than anything previously
attempted whilst evolving far bigger programs.

As expected,
without size or depth limits or biases
crossover with brutal selection pressure
tends to evolve very large non-parsimonious programs.
Known in the GP community as bloat
(page~\pageref{p.bloat} \cite[page~617]{koza:book}).
After a few initial generations,
GP tree bloat typically follows a
sub-quadratic power law~\cite{langdon:2000:quad}.
But eventually effective selection pressure 
\cite[sec.~14.2]{nordin:thesis},
\cite[page~187]{banzhaf:1997:book},
\cite{Stephens:1999:ECJ},
\cite{langdon:fogp}
within highly evolved populations falls,
leading to bloat at a reduced rate.
However we only see the chaotic lack of bloat found in long running
Boolean problems
\cite{Langdon:2017:GECCO}
in a few unsuccessful runs with tiny populations
(red plots in Figure~\ref{fig:sextic_p48}, 
page~\pageref{fig:sextic_p48}).
Nevertheless in all cases bloated binary trees
evolve to be randomly shaped and lie close to Flajolet's 
square root limit
(Section~\ref{p.Flajolet}).

Evolving
binary Sextic polynomial trees for up to a million generations,
during which some programs grow to four hundred million nodes,
suggests 
even a simple floating point benchmark allows 
long term fitness improvement over thousands of generations.

\section*{Acknowledgements}
This work was inspired by conversations
at Dagstuhl 
\href{https://www.dagstuhl.de/en/program/calendar/semhp/?semnr=18052}
{Seminar 18052} on
Genetic Improvement of Software.

The new parallel GPQuick code is
available via 
\href{http://www.cs.ucl.ac.uk/staff/W.Langdon/ftp/gp-code/GPavx.tar.gz}
{\tt  http://\allowbreak{}www.\allowbreak{}cs.\allowbreak{}ucl.\allowbreak{}ac.\allowbreak{}uk/\allowbreak{}staff/\allowbreak{}W.Langdon/\allowbreak{}ftp/gp-code/\allowbreak{}GPavx.tar.gz}

\footnotesize
\bibliographystyle{apalike}
\bibliography{/tmp/gp-bibliography,/tmp/references} 

\end{document}